\documentclass[journal]{IEEEtran}
\usepackage{color}
\usepackage{times}
\usepackage{epsfig}
\usepackage{graphicx}
\usepackage{amsmath}
\usepackage{amssymb}
\usepackage{multirow}
\usepackage{floatrow}
\usepackage{mathrsfs}
\usepackage{array}
\usepackage{ifpdf}
\usepackage{algorithmic}
\usepackage{algorithm}
\usepackage{cite}
\usepackage{subcaption}
\usepackage{changes}

\makeatletter
\let\Changes@Markup@Deleted\@gobble
\makeatother
\usepackage{url}
\usepackage[colorlinks,
            linkcolor=red,
            anchorcolor=blue,
            citecolor=green 
            ]{hyperref}

\begin{document}
\title{OccluMix: Towards De-Occlusion Virtual Try-on by Semantically-Guided Mixup}
\author{ Zhijing Yang\textsuperscript{\dag}, Junyang Chen\textsuperscript{\dag}, Yukai Shi\IEEEauthorrefmark{1}, Hao Li, Tianshui Chen, Liang Lin

\thanks{
 {\dag} The first two authors share equal contribution.
 }
 
\thanks{
  Code is available at: https://github.com/JyChen9811/DOC-VTON
 }
 
}

\maketitle

\begin{abstract}
Image Virtual try-on aims at replacing the cloth on a personal image with a garment image (in-shop clothes), which has attracted increasing attention from the multimedia and computer vision communities. Prior methods successfully preserve the character of clothing images, however, occlusion remains a pernicious effect for realistic virtual try-on. In this work, we first present a comprehensive analysis of the occlusions and categorize them into two aspects: i) Inherent-Occlusion: the ghost of the former cloth still exists in the try-on image; ii) Acquired-Occlusion: the target cloth warps to the unreasonable body part. Based on the in-depth analysis, we find that the occlusions can be simulated by a novel semantically-guided mixup module, which can generate semantic-specific occluded images that work together with the try-on images to facilitate training a de-occlusion try-on (DOC-VTON) framework. Specifically, DOC-VTON first conducts a sharpened semantic parsing on the try-on person. Aided by semantics guidance and pose prior, various complexities of texture are selectively blending with human parts in a copy-and-paste manner. Then, the Generative Module (GM) is utilized to take charge of synthesizing the final try-on image and learning to de-occlusion jointly. In comparison to the state-of-the-art methods, DOC-VTON achieves better perceptual quality by reducing occlusion effects. 
\end{abstract}


%
\IEEEpeerreviewmaketitle

\section{Introduction}
\IEEEPARstart{V}{irtual} try-on is a popular application by transferring a desired in-shop clothing onto a reference person. With the demand of e-business, virtual try-on has been attracted rising attention. Although recent developments in the try-on network have helped to perform realistic cloth warping by generating fitness shape and realistic visual quality try-on images \cite{Viton,ACGPN,clothflow,HDviton,CT-Net,disentangled,PFAFN}, it remains a big challenge to locate and resolve the occlusion effect in the distorted try-on image.

\begin{figure}[!t]
\centering
\includegraphics[width=1.0\linewidth]{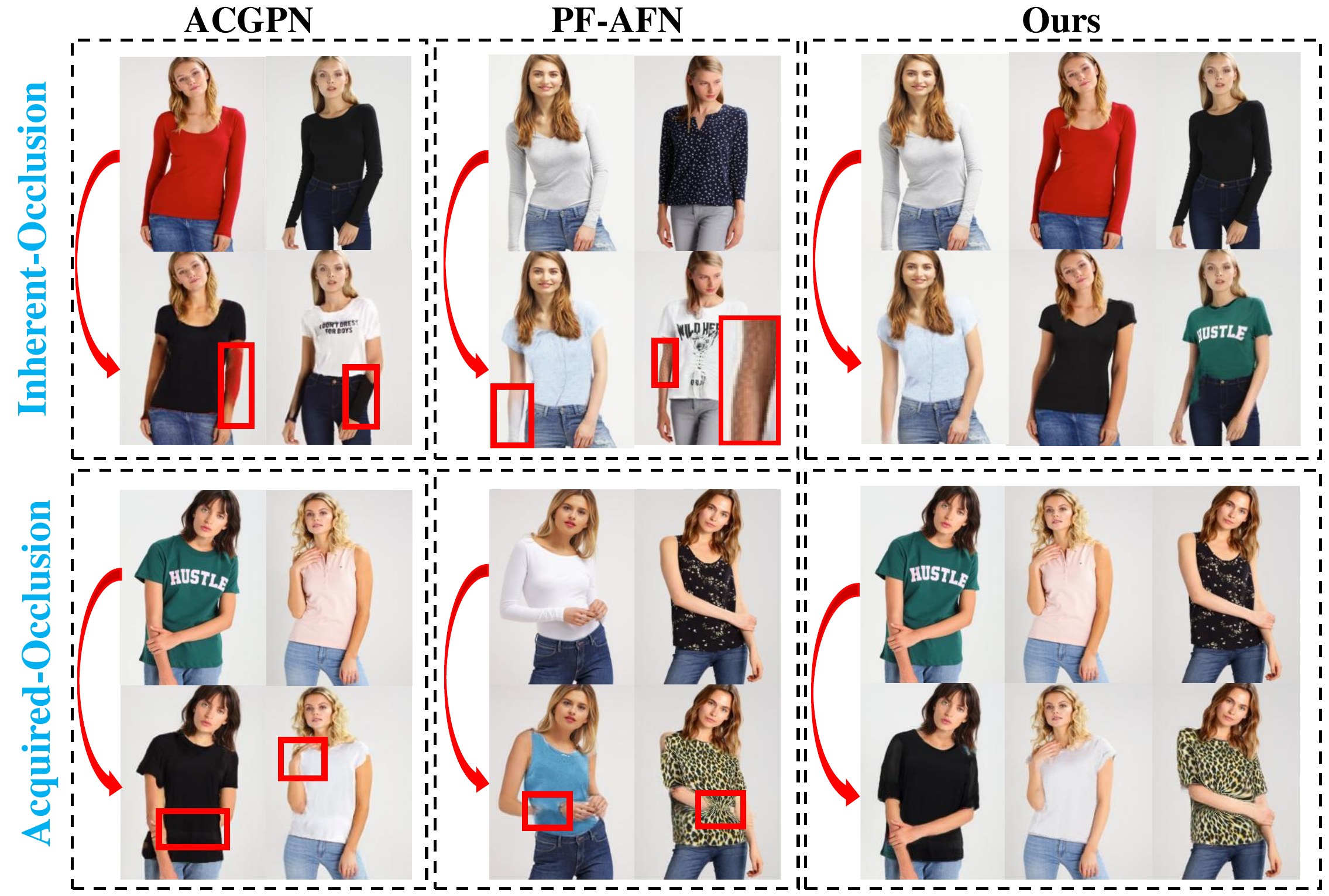}
\caption{On Viton~\cite{Viton} dataset, the try-on results of the state-of-the-art models~\cite{ACGPN,PFAFN} appear undesired occlusion from former and new clothes. To address the occlusion phenomenon, we attempt to categorize the majority of occlusion into two types (\textit{i.e.,} Inherent-Occlusion and Acquired-Occlusion). In the supplementary file, we provide massive occlusion samples to verify Inherent- and Acquired- types.} 
\label{introduction_image}
\vspace{-4mm}
\end{figure}

\begin{figure*}[t]
    \centering
 \vspace{1mm}
 \begin{subfigure}{0.45\textwidth}
    \includegraphics[width=\textwidth]{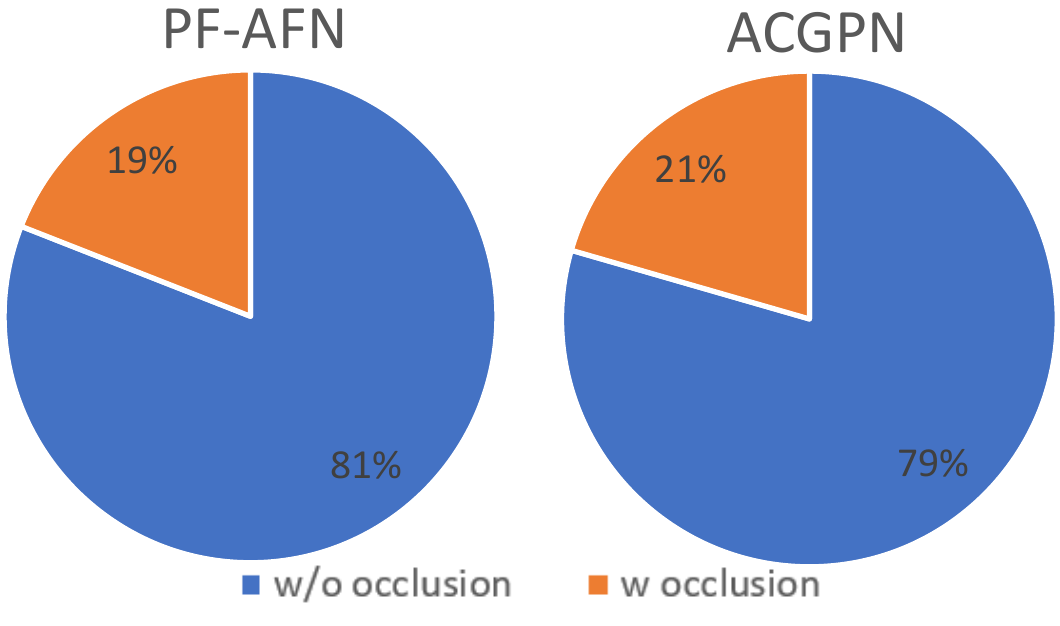}
    \caption{}
 \end{subfigure}\hspace{6mm}
 \begin{subfigure}{0.45\textwidth}
    \includegraphics[width=\textwidth]{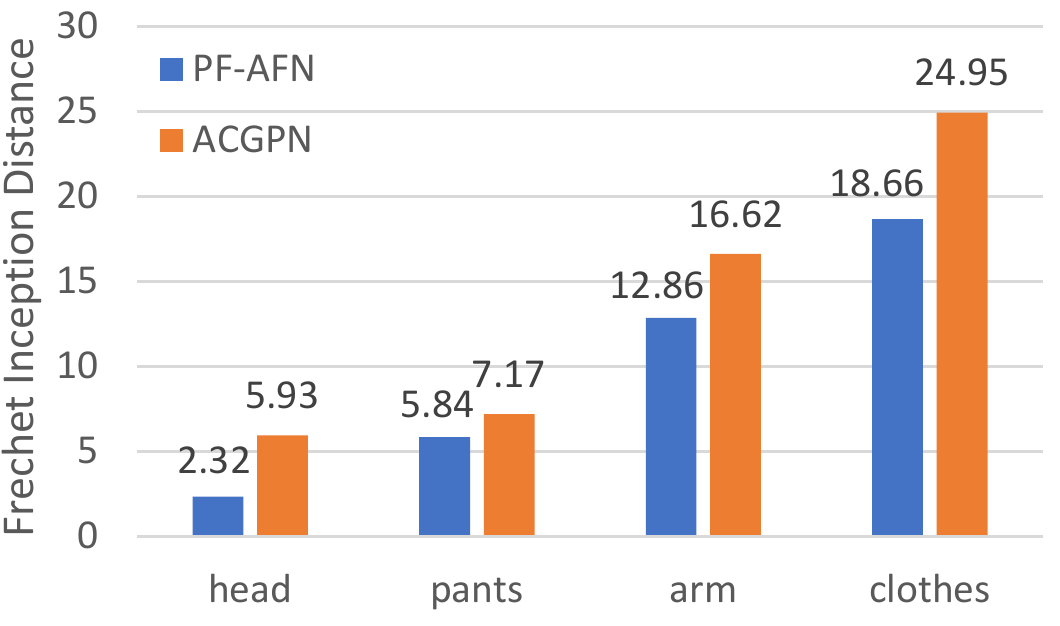}
    \caption{}
 \end{subfigure}
 \vspace{2mm}
 \caption{(a) To investigate the proportion of occlusion samples, we count occlusion examples of PF-AFN and ACGPN results on CP-VTON+~\cite{CP_VTON_plus} dataset aided by human annotation. (b) The Frechet Inception Distance (FID) scores of different body parts generated by PF-AFN~\cite{PFAFN} and ACGPN~\cite{ACGPN}.}
   \label{fig:motivation}
\vspace{-6mm}
\end{figure*}

To investigate the occlusion effect in existing visual try-on methods, we follow the pipeline of CP-VTON+~\cite{CP_VTON_plus} to explore the results of two representative methods~\cite{ACGPN,PFAFN}. In our experiments, ACGPN~\cite{ACGPN} stands for the parser-based methods and PF-AFN~\cite{PFAFN} stands for the parser-free methods. As shown in Fig.~\ref{introduction_image}, in the first and the second rows, the occlusion in the try-on image is represented as the ghost of the previous garment. We denote this occlusion as Inherent-Occlusion. Typically, inherent-occlusion is caused by poor generalization and wrong human parsing. As shown in the third and the fourth rows in Fig.~\ref{introduction_image}, the occlusion effect is caused by the wrong shape warping of new clothes. We denote it as Acquired-Occlusion. Hence, we mainly demonstrate the occlusion problems into two forms (\textit{i.e.,} Inherent- and Acquired- Occlusion). The cloth warping module in the try-on workflow is easily misled by spatial transformation, and the occlusion effect becomes an obvious degradation when the human pose exhibits large variance. To this end, some pioneering synthesized-based models~\cite{Viton,clothflow,cp-Vton,wuton} lock into the above drawbacks and demonstrate limited image quality.

In Fig.~\ref{fig:motivation}, we conducted a human annotation to investigate the occlusion effect in virtual try-on methods. Specifically, 4064 images~\cite{CP_VTON_plus} are labeled with the existence as well as location of occlusion by manual effort. The statistical results show that occlusion is still the main challenge in the try-on task with 20\% percent occurrence. We further calculate the mean Frechet Inception Distance (FID) value of each body part in the results. The arm and clothing parts account for a large FID score and spatial variance, which motivates us to improve the image quality by focusing on these challenging human parts.

To tackle the aforementioned issues, we present a robust De-OCclusion framework for Virtual Try-on (DOC-VTON), which fully applies the semantic layout of the try-on image. DOC-VTON adaptively performs a crop-and-paste operation~\cite{cutmix,mixup} for the generative module (GM) to implement de-occlusion. Specifically, the DOC-VTON consists of three modules: i) Cloth Warping Module (CWM), which warps in-shop clothing and its corresponding mask into the fitting shape by using appearance flow; ii) Occlusion Mixup Module (OccluMix), which simulates different occlusion cases as OccluMix samples based on the semantic layout of Sharpened Parsing Network (SPN); iii) Generative Module (GM) is applied to transfer the clothes on the OccluMix sample to the real person image, enabling final try-on image generation and de-occlusion jointly.
 
With the above modules, DOC-VTON learns de-occlusion by adopting a semantically-guided mixup strategy in virtual try-on. In summary, the main contributions of our paper are as follows:

\begin{itemize}

\item We present a comprehensive analysis of the occlusion effect of current visual try-on algorithms. This is the first attempt to analyze this point, and it can facilitate further research on de-occlusion virtual try-on.

\item We propose a simple yet non-trivial occlusion mixup strategy for virtual try-on (OccluMix), which obtains the challenging occluded try-on person by blending various complexity of texture with semantic and posture guidance.

\item Compared with the general parsing pipeline, we investigate a Sharpened Parsing Network (SPN) to parse try-on images iteratively. SPN not only handles the irrational warping parts surgically, but also affords regions for OccluMix.

\item Extensive experiments and evaluations demonstrate that our method can achieve the best state-of-the-art results in the VITON task qualitatively and quantitatively.
\end{itemize}

In the remaining parts of this paper, Section II briefly surveys the occlusion phenomenon in existing virtual try-on approaches, and the derivatives of data augmentation. Section III presents a comprehensive occlusion analysis of state-of-the-art methods. Section IV presents DOC-VTON pipeline and detailed explanations of OccluMix.
In Section V, we discuss the comparison between OccluMix and other data augmentation methods. In Section VI, we perform experiments to verify the effectiveness and efficiency of DOC-VTON by comparing it with existing state-of-the-arts. Finally, we conclude our work with future research directions in Section VII.



\begin{figure*}[t]
	\begin{center}
		\includegraphics[width=0.98\linewidth]{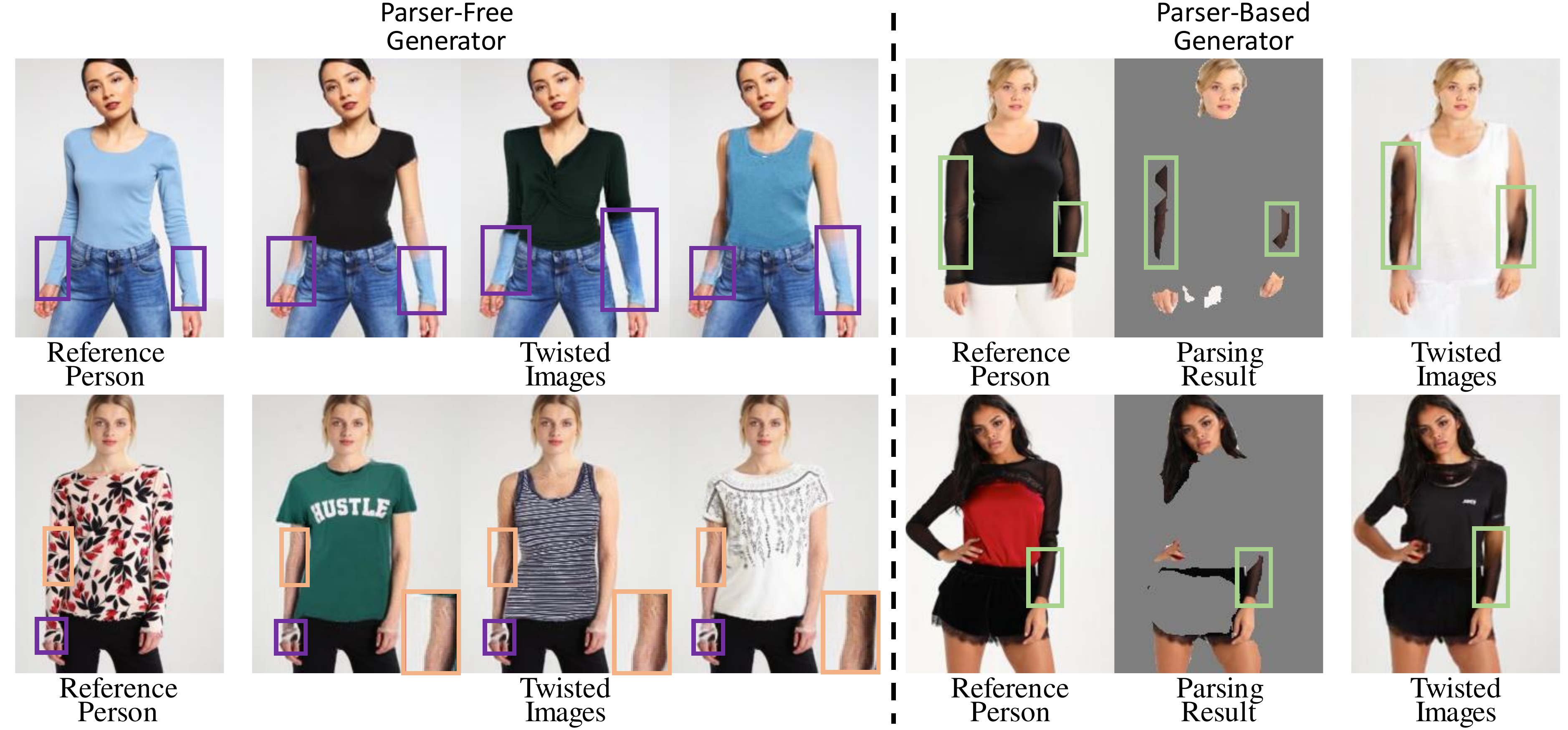}
	\end{center}
	\vspace{0mm}
	\caption{Analysis of Inherent-Occlusion. For parser-free try-on results, the ghost of previous garment will remain no matter try on arbitrary garments. Compared to parser-free method, the parser-based method will generate twisted images with failed parsing results.}
	\label{fig:inherent}
	\vspace{-6mm}
\end{figure*}
\section{Related Work}
\textbf{Virtual Try-on}. Existing deep learning-based methods on virtual try-on can be mainly categorized as 3D model based approaches~\cite{3d_1,3d_2,3d_3,3d_4} and 2D image-based ones~\cite{Viton,ACGPN,PFAFN,cp-Vton,2d_multil-pose,SPG_VTON,Vton-scfa}. As the former methods require 3D measurements, which bring extra computation resources. Instead, 2D image-based approaches are more feasible to real-world scenarios. For example, VITON~\cite{Viton}, CP-VTON~\cite{cp-Vton}, ACGPN~\cite{ACGPN}, ClothFlow~\cite{clothflow}, DCTON~\cite{DCTON} and RT-VTON~\cite{RT_VTON} use human representation as the input to generate a clothed person. Besides, WUTON~\cite{wuton} and PF-AFN~\cite{PFAFN} employ a parser-free approach. In garment deformation methods, the VITON~\cite{Viton}, CP-VTON~\cite{cp-Vton}, ACGPN~\cite{ACGPN}, DCTON~\cite{DCTON} and VITON-HD~\cite{VITON_HD} use thin-plate-spline (TPS)~\cite{tps} transformation to warp target cloth into fitness shape. However, TPS transformation exhibits limited deformation ability. To this end, RT-VTON~\cite{RT_VTON} proposed a semi-rigid deformation to align the warped cloth with the predicted semantics. ClothFlow~\cite{clothflow} and PF-AFN~\cite{PFAFN} use appearance flow~\cite{appearance_flow}, which easily warps the target clothes smoothly onto the target person.

As typical try-on pipelines, VITON and CP-VTON use rough shapes and pose maps to ensure the generalization of arbitrary clothes. However, parser-based methods~\cite{Viton,ACGPN,cp-Vton,Vtnfp} generate poor quality try-on images when parsing results become inaccurate. Recently, PFAFN~\cite{PFAFN} proposes a pioneering parser-free knowledge distilling approach that gets rid of the interference from inaccurate segmentation. Nevertheless, the generated images of PFAFN still encounter the Inherent-Occlusion and Acquired-Occlusion. To this end, we propose a novel De-occlusion method for virtual try-on (DOC-VTON) . DOC-VTON handles the misalignment between target clothes and the reference person, and reduces the ghost effect caused by previous clothes.

\textbf{De-Occlusion}. Removing the partial occlusion from the target object is a crucial computer vision task~\cite{occulusion, 2019Iterative, amodal_18, zhang2022face, dong2016occlusion, Occlusion-Handling}. Sail~\cite{hu2019sail} proposes a novel self-supervised framework that tackles scene de-occlusion on real-world data without manual annotations. GAN-based~\cite{zhang2022face} methods are used to inpaint the occluded region on the face. In the virtual try-on task, the occlusion problem also exists by covering the clothes and human body up. To avoid the occlusion phenomenon, the parser-based methods~\cite{ Viton, ACGPN,cp-Vton, Vtnfp} use a human parser to understand the spatial layout of the body part. Nevertheless, parser-based methods tend to generate poor quality try-on images with noticeable occlusion. Recently, PFAFN~\cite{PFAFN} proposes a pioneering parser-free approach, however, the image quality still suffers from unsuitable shape-warping clothes and general model generation capabilities. To address the above problems, we introduce the OccluMix strategy into the Virtual Try-On task.

\textbf{Data augmentation}. Recently, several studies employ data augmentation (DA)~\cite{DA10, stylemix, cutblur,  supermix, Vehicle, kang2021data, random_crop} to enhance generalization ability. Mixup~\cite{mixup, DA9} uses convex linear interpolation on the image level for data augmentation. CutMix~\cite{cutmix} proposes to cut and paste a cropped area from an input image to other images for data augmentation. Although MixUp and CutMix demonstrate practical improvements, they do not utilize image prior knowledge, such as saliency, semantics and optical flow, for guidance. Random Crop~\cite{random_crop} crops the images into a particular dimension and creates synthetic data. Compared with Mixup and Cutmix, it flexibly preserves the prior knowledge and solves the distortion problem caused by different scale try-on images~\cite{kang2021data}. However, it still remains a big challenge to solve the occlusion phenomenon (\textit{i.e.,} Inherent- and Acquired- Occlusion) on the simple scale image. Inspired by Supermix~\cite{supermix}, we investigate a goal-oriented data augmentation method by using human parsing priors~\cite{Parsing} for the try-on generation model.

\begin{figure}[t]
	\includegraphics[width=0.98\linewidth,height=0.8\linewidth]{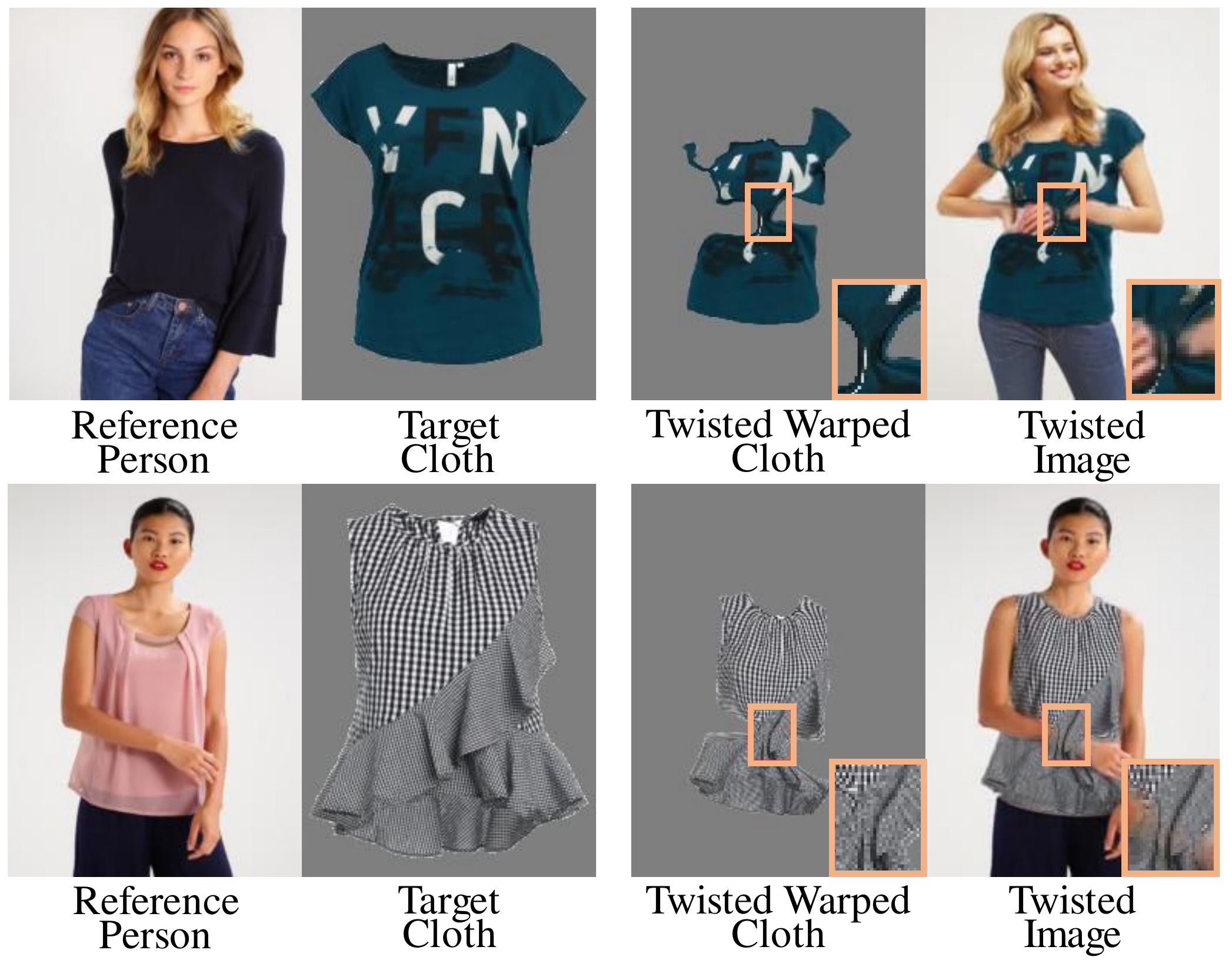}
	\vspace{3mm}
	\caption{Try-on images inherit the twisted pattern of the twisted warped cloth.}
	\label{fig:acquired}
	\vspace{-6mm}
\end{figure}

\begin{figure*}[t]
	\begin{center}
		\includegraphics[width=0.95\linewidth]{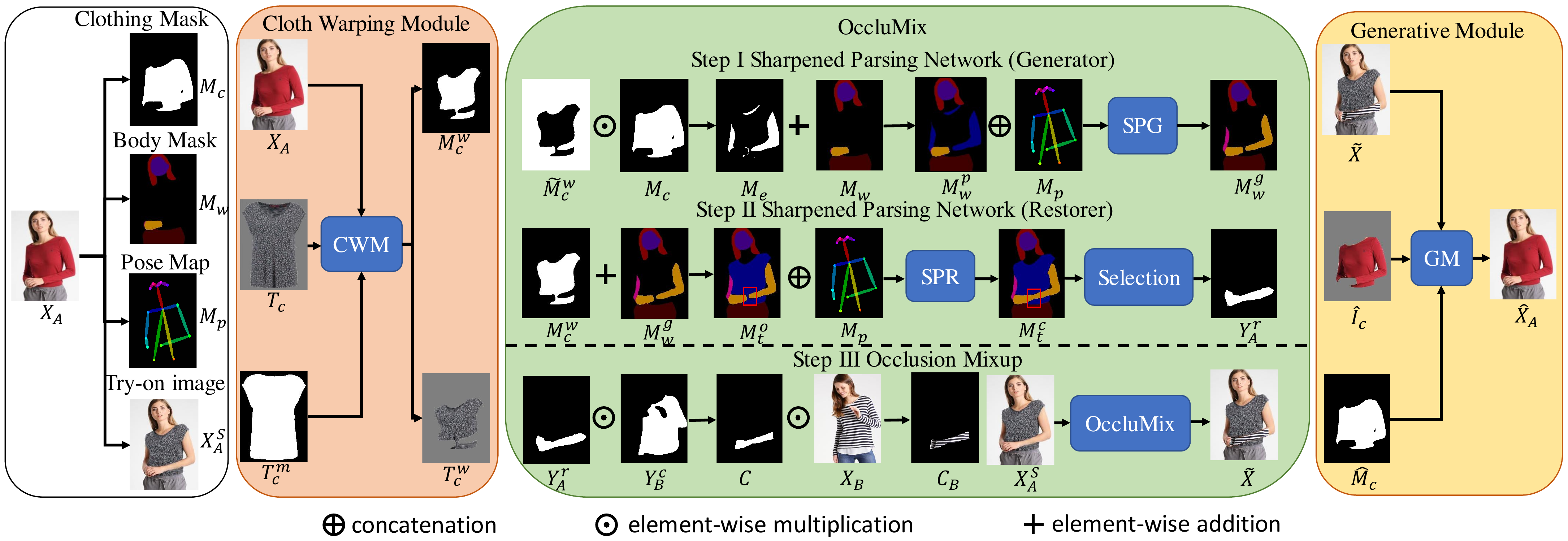}
	\end{center}
	\vspace{2mm}
	\caption{An illustration of our proposed method. OccluMix generates an augmented image by crop-and-pasting different clothing textures onto the challenging region of the try-on image. To simulate a challenging occlusion emergence on the try-on image, we need to identify the human component of the try-on image. We first obtain parser-based input image $X_A$, then use the Cloth Warping Module (CWM) to predict the warping cloth $T_C^W$ and warping clothing mask $M_C^W$; In Step I, the Sharpened Parsing Generator ($G$) first multiplies the Clothing Mask $M_C$ and $\tilde{M}_C^W$ to get the strange area $M_e$, then combine it with the body parts $M_w$ to produce the potential location of body parts $M_w^p$ (including head, arms, and pants). Then, $G$ generates the rough mask of body parts $M_w^g$ by using Pose Map $M_p$ and the potential location of body parts $M_w^p$; In step II, the Sharpened Parsing Restorer ($R$) refines the rough mask of torso $M_t^o$ to get the complete torso mask $M_t^c$. In step III, we use DensePose~\cite{densepose} to select a challenging segment $Y_A^r$ to multiply with an auxiliary cloth $X_B$ to obtain the texture occlusion $C_B$. And we mix $C_B$ with the try-on image $X_A^s$ to get the OccluMix sample $\tilde{X}$. Finally, we exploit a generative module to generate the try-on images $\hat{X}_A$ by utilizing the warped cloth information $\hat{I}_c$,$\hat{M}_c$ and $\tilde{X}$.}
	\label{fig:network}
	\vspace{-5mm}
\end{figure*}

\section{Occlusion Analysis}
Existing state-of-the-art methods on 2D virtual try-on can be classified as a parser-based approach and a parser-free approach. However, both of them still remain the occlusion phenomenon in the try-on images. To investigate the occlusion effects in both approaches, we follow the pipeline of CP-VTON+ to explore the occlusion phenomenon on ACGPN and PF-AFN. 
As the former represents the parser-based approach, and PF-AFN represents the parser-free approach. We first explore the occlusion effect caused by twisted warped cloth. As shown in Fig.~\ref{fig:acquired}, if the warped results remain twisted pattern, the generated results will remain in this pattern, and express it in the form of occlusion. We denoted this occlusion as acquired-occlusion, which was caused by the twisted warped results. It seems that the generator will generate clean try-on images with proper warping results, however, the occlusion effect also exists in the generative stage. As shown in Fig.~\ref{fig:inherent}, try-on results will remain the ghost of the previous garment. For a parser-free generator, we find that the ghost will remain no matter the try-on arbitrary garments. Since the parser-free generator needs to classify the try-on region of the reference images, we assume that the ghost is caused by poor generalization ability. In comparison with the parser-free generator, the parser-based generator will generate twisted images with failed parsing results. We denote this occlusion effect caused by previous garments as inherent-occlusion. Hence, we mainly demonstrate the occlusion problems in two forms. (\emph{i.e.}, Inherent- and Acquired- Occlusion).

\begin{figure*}[ht]
	\begin{center}
		\includegraphics[width=0.98\linewidth]{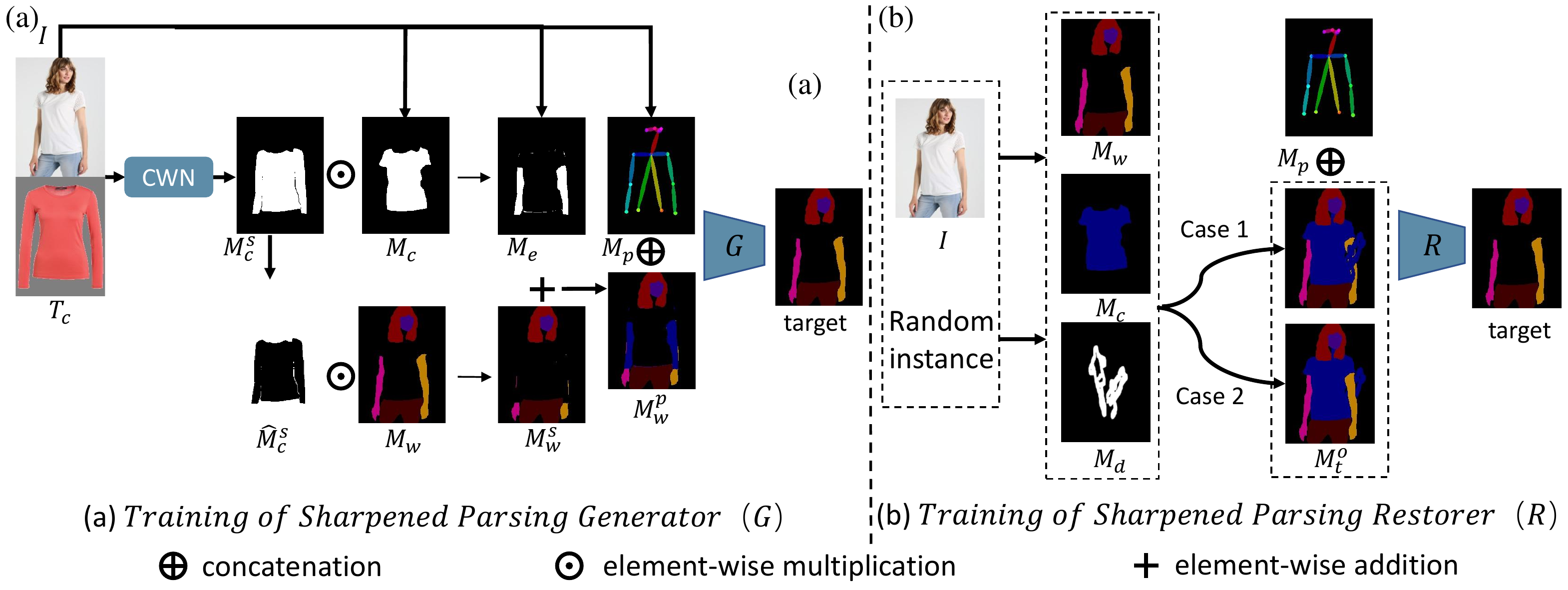}
	\end{center}
	\vspace{2mm}
	\caption{(a) Generator is fed with Pose Map $M_p$ and the potential location $M_w^p$ to generate the coarse mask of body. (b) Restorer is trained to refine the body mask. At this stage, we divide two training cases. Case 1 is encouraged to partially complete the mask of body parts. Case 2 prevents Restorer from over-completing.}
	\label{fig:self_sup}
	\vspace{-5mm}
\end{figure*}

\section{Methodology}
The proposed DOC-VTON is composed of three modules, as shown in Fig.~\ref{fig:network}. First, the Clothes Warping Module is designed to warp the target clothing image onto a real image. Second, the OccluMix module progressively generates the mask of body parts of the try-on image via semantic information, yielding occlusion effect by utilizing crop-and-paste strategy on try-on image to generate OccluMix samples. Finally, the generative module synthesizes the try-on image.

\subsection{Cloth Warping Module (CWM)}

Following the training pipeline of PF-AFN~\cite{PFAFN}, we use the second-order constraint to better preserve the cloth characteristics, and the constraint is defined as follows:
\begin{equation}
L_{sec} = \sum ^ N _{i=1} \sum_p \sum_{\pi \in N_p}Char(f^{p-\pi}_i + f^{p+\pi}_i - 2f^p_i),
\vspace{-3mm}
\end{equation}
where $f^p_i$ denotes the p-th point on the flow maps of i-th scale. $Char$ is the generalized charbonnier loss function~\cite{charbonnier_loss}. $N_p$ consists of the set of vertical, horizontal, and both diagonal neighborhoods around the p-th point. 

CWM warps the target cloth into a fit shape, which also maintains the details of the cloth. It performs well when the reference person stands simply. When a reference person stands in a complex posture, such as both torso twisting and two hands blocking in front of the body, the warping cloth may cover part of the human body.

\subsection{OccluMix}
\subsubsection{Sharpened Parsing Network (SPN)}

The sharpened parsing network (SPN) is proposed to refine the warping clothes as well as to generate the body parts (e.g., arms) of the person. Many previous works neglect the fact that accurate parsing results can correct the unreasonable warping process. To address this issue, a sharpened parsing mechanism is adopted to refine the detail distortion of human parsing during try-on transformation.

Specifically, suppose a person tries on new clothes, we define the masks of clothing and body parts as $M_{c}$ and $M_{w}$ (including head, arms and pants) in the original image, $M_{c}^s$ and $M_{w}^s$ are the masks of clothing and body parts in the try-on image, and $M_{p}$ is the skeleton information of the reference person. Since $M_{w}^s$ is absent, we use cloth and torso prior with warping cloth to obtain it. The formulation of $M_{w}^s$ is defined as follows:

\begin{equation}
    M_{w}^s = M_{w} \odot (1-M_{c}^s).
\end{equation}
where $\odot$ indicates element-wise multiplication. And $M_{e}$ is the mask of strange fabric between $M_{c}^s$ and $M_{c}$, which indicates the potential location for generating torso region. The formulation of $M_{e}$ is:
\begin{equation}
    M_e = M_{c}^s \odot (1-M_{c}).
\end{equation}
Then, we get the potential location of body parts $M_{w}^p$ in try-on image by combining $M_{e}$ and $M_{w}^s$ as follows:

\begin{equation}
M_{w}^p = M_{w}^s + M_{e}.
\end{equation}

As shown in Fig.~\ref{fig:self_sup} (a), since $M_{w}^p$ and $M_{w}$ are obtained, we can use the process of $(M_{p},M_{w}^p)$ ${\xrightarrow{G}}M_{w}$ to realize $G$ to generate the rest masks of torso from the potential location $M_{w}^p$ under the supervision of the label $M_{w}$.

However, when the reference people stand in a twisted posture, the pernicious distortion of the target cloth may destroy the body details in $M_w$. To address this problem, we introduce $M_{d}$ from the Irregular Mask Dataset~\cite{liu2018image} and merge it with $M_c$ to simulate the failure case $M_t^o$, where some details of the body are lost. The formulation is defined as follows:
\begin{equation}
{M_t^o} = M_{w} \odot (1 - M_{d}) + M_{c}\cup M_{d}.
\label{gonshi1}
\end{equation}

We then feed it into restorer $R$, and $(M_t^o, M_{p})$ ${\xrightarrow{R}}M_w$ is the process to refine the details of body parts. Note that training details of the restorer are presented in Fig.~\ref{fig:self_sup} (b).

\begin{figure}[ht]
\centering
\includegraphics[width=0.95\linewidth]{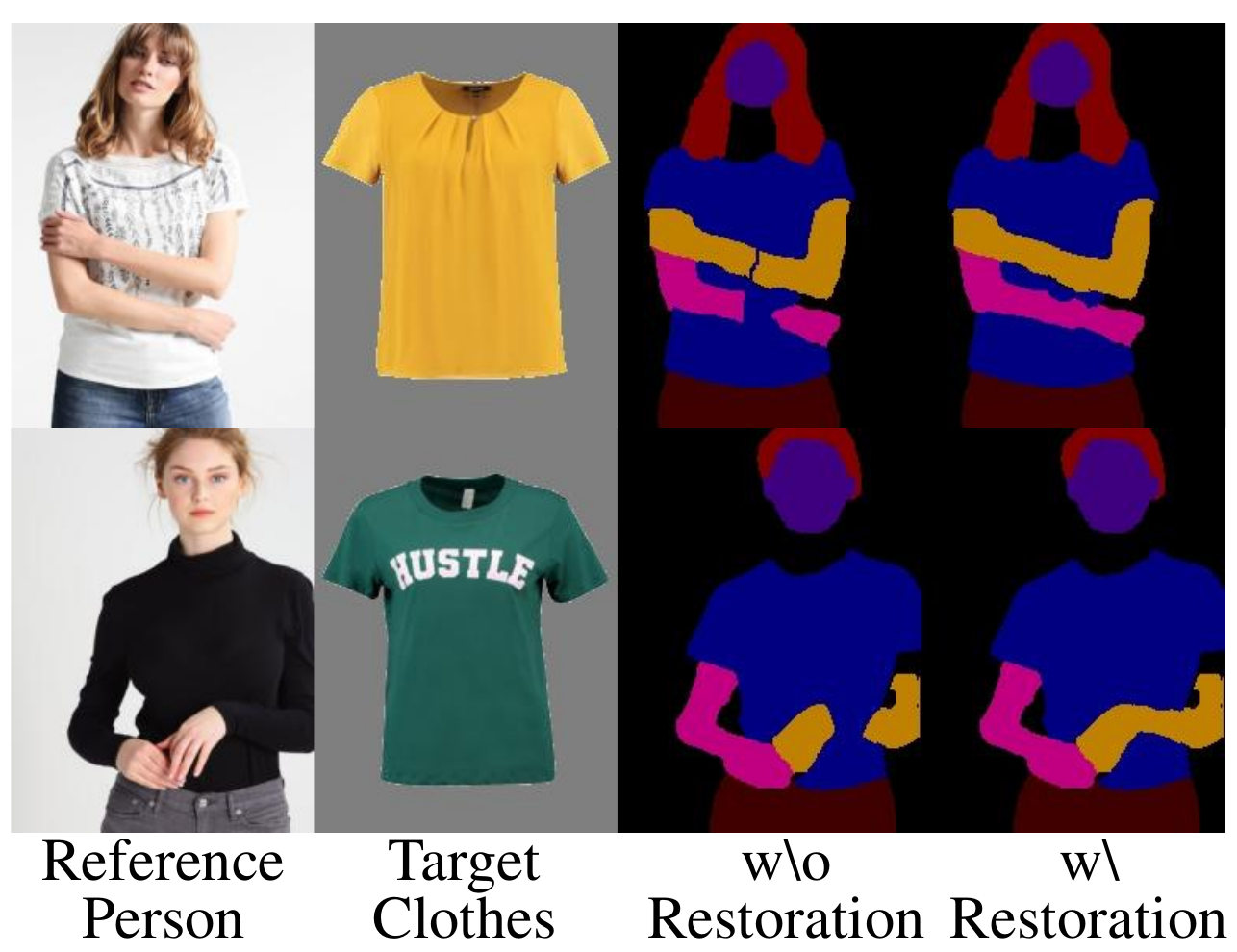}
\vspace{2mm}
\caption{Effect of semantic restoration component. When the reference people stands in a twisted posture, Sharpened Parsing Generator provides wrong semantic information on try-on images, the restoration step can fixed this question.}
\label{fig:self_sup_effect}
\vspace{-3mm}
\end{figure}

As shown in Fig.~\ref{fig:self_sup_effect}, the detail distortion of the body parts in $M_o$ is enhanced by Restorer. After refining the parsing mask of the try-on image, we can use it to mine the regions to crop-and-paste the texture occlusion, and model the OccluMix data.


\subsubsection{Occlusion Mixup}

In this section, we describe OccluMix, a data augmentation (DA) strategy that is designed for the try-on task. A practical DA method for try-on needs to simulate the challenging occlusion and serve as a good regularizer for the try-on model. Literally, OccluMix overlays different clothes on the try-on images, enforcing the network to restore realistic details during the try-on transformation.

\begin{figure}[ht]
\centering
\includegraphics[width=0.95\linewidth]{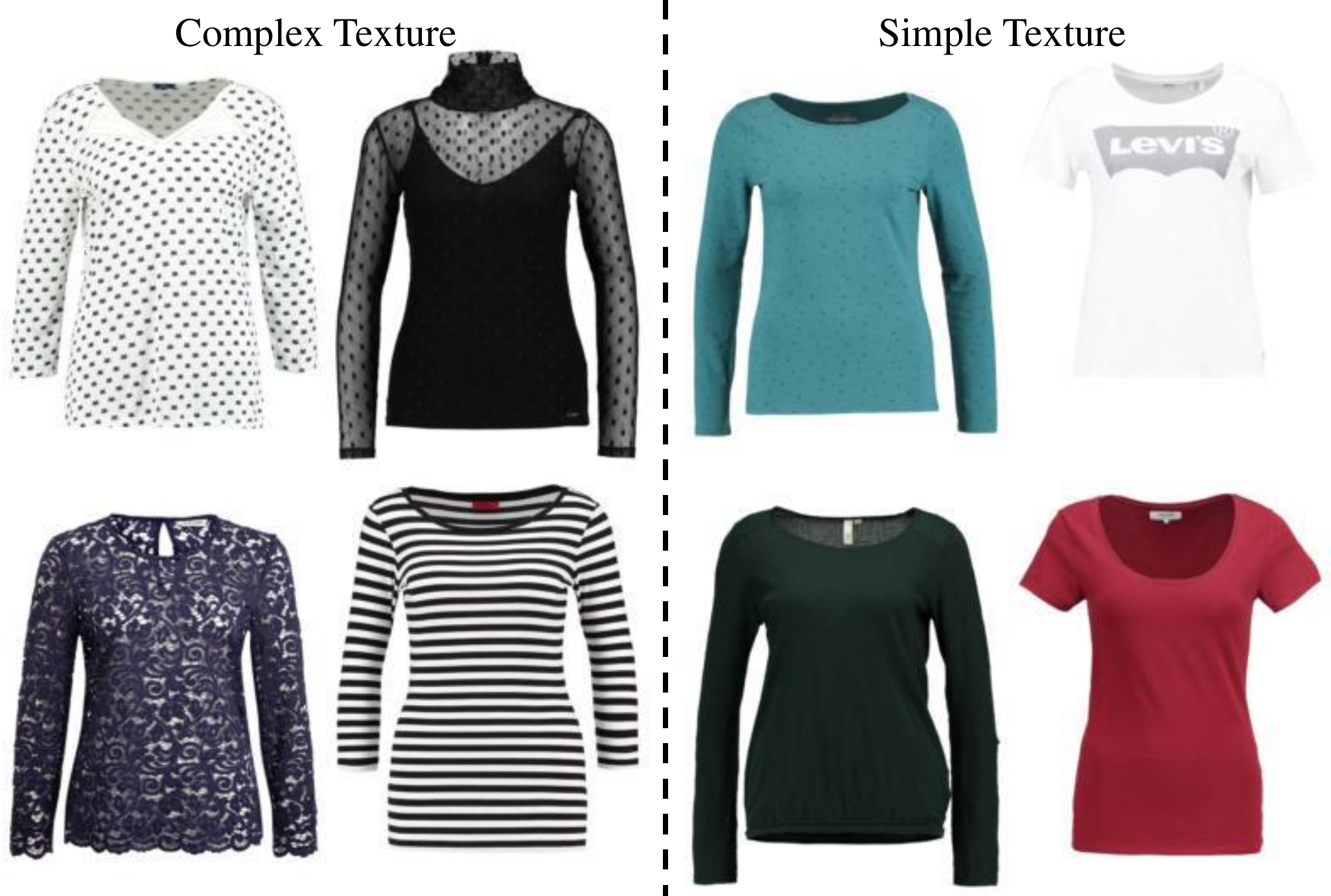}
\vspace{0mm}
\caption{The left clothes are complicated by involving dots, stripes, and various other textures. And the right clothes are more simple with less texture.}
\label{fig:cloth_category}
\vspace{-3mm}
\end{figure}
In the top right of Fig.~\ref{introduction_image}, the personal image 
will retain the ghost of the complex texture when a person wearing complex clothes wants to put on new cloth. To this end, we ensure a certain percentage of complex textures existing in OccluMix. In our experiment, we use Gray-Level Co-occurrence Matrix~\cite{gray_rock} to estimate the entropy of the clothing complexity. Besides, we divide the clothes into two categories (\textit{i.e.,} simple and complex) based on their texture complexities:
\begin{equation}
ENT = -\sum^k_{i=1} \sum^k_{j=1} G(i,j)logG(i,j),
\end{equation}
where $G(i,j)$ represents the normalized occurrence of different gray scale values.

As shown in Fig.~\ref{fig:cloth_category}, the left clothes are categorized into complex textures, and the right are simple textures. Then we use texture categorization to divide the training pairs of the reference person.

\begin{equation}
T=
\left\{\begin{array}{l}
1,~ENT>=2.5,\\ 
0,~ENT<2.5,
\end{array}\right.
\end{equation}
where $T = 1$ indicates that the clothes belong to the complex texture category and the other to the simple texture category.



Let $x\in \mathbb{R}^{W\times H\times C}$ and $y$ denote a random sampled image and corresponding label,
respectively. Example $A$ (\textit{i.e.,} $(x_A,y_A)$) is the training sample. And example $B$ (\textit{i.e.,} $(x_B,y_B)$) is chosen from the complex texture set or the simple texture set by complex coefficient $\lambda$. 

\begin{figure}[t]
	\includegraphics[width=0.88\linewidth,height=0.8\linewidth]{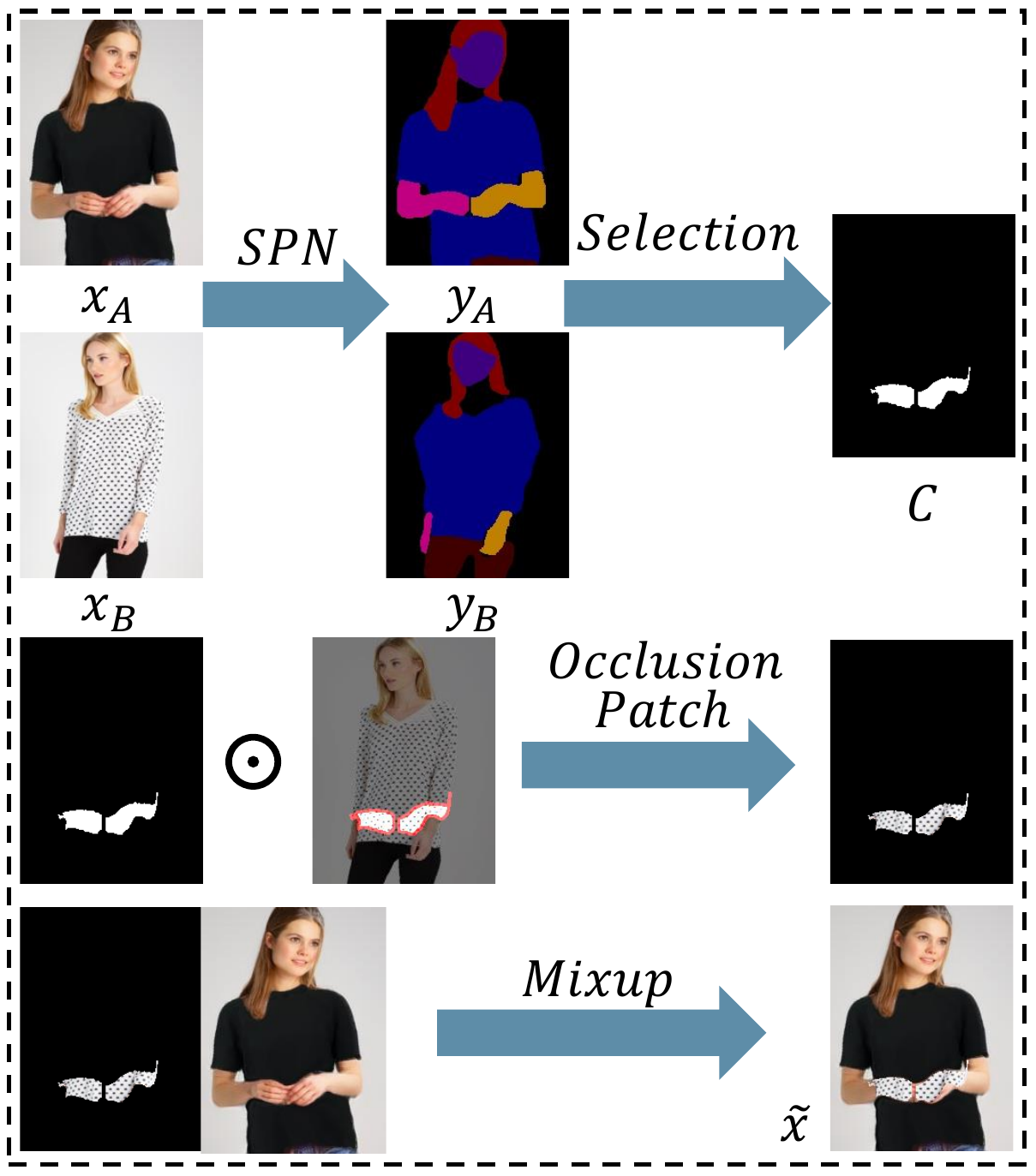}
	\vspace{-2mm}
	\caption{Process of Occlusion Mixup. We attempt to model occlusion in try-on images, however, it is hard to enforce texture occlusion to be gathered in a typical region to model real occlusion. To tackle this challenge, we utilize Sharpened Parsing Network (SPN) in OccluMix.}
	\label{fig:process_of_cutmix}
	\vspace{0mm}
\end{figure}

As shown in Fig.~\ref{fig:process_of_cutmix}, the goal of try-on mixup is to generate a new training sample $\tilde{x}$ by combining two training samples $(x_A,y_A)$ and $(x_B,y_B)$. The formulations are as follows:
\begin{equation}
\begin{split}
    &C = y_A^{r} \odot y^{c}_B,\\
    \vspace{3mm}
\tilde{x} = C &\odot x_B + (1-C) \odot x_A,
\end{split}
\end{equation}
where $y_A^{r}$ is the selected region for occlusion mixing. We first count the \textit{occlusion distribution} of each human part in try-on images. DensePose~\cite{densepose} is then used to forward the results of Sharpened Parsing Network, and select $y_A^{r}$ out w.r.t the \textit{occlusion distribution}. Besides, $y_B^{c}$ is the clothing mask of the person $x_B$ and $C$ is the mask of the texture occlusion. As shown in Fig.~\ref{fig:network}, we mix the cropped clothes into the input data to generate the augmented images. Finally, we feed the augmented images into the generator to obtain the try-on images with clothes on the real images.

\subsection{Generative Module}
In this module, we adopt Res-UNet~\cite{clothflow} as the backbone architecture of the generative module (GM). It can not only retain the characteristics of warped clothes, but also keep the details of human body parts. 

In the training phase, the parameters of GM are optimized by minimizing $L_g$, as follows:
\begin{equation}
L_g = \alpha_l L_{euc} + \alpha_p L_{per},
\end{equation}
where $L_{euc}$ is the pixel-wise L1 loss and $L_{per}$ is the perceptual loss~\cite{VGGloss} to encourage the improvement of the try-on image visual quality. The formulations are as below: 

\begin{equation}
L_{euc} = \lVert I^G - I\rVert_1,
\end{equation}
\vspace{-4mm}
\begin{equation}
L_{per} = \sum_m \lVert \phi_m(I^G) - \phi_m(I)\rVert_1, 
\end{equation}
where $I^G$ and $I$ are the generated and real image, respectively. And $\phi_m$ indicates the $m$-th feature map in a VGG-19~\cite{VGG} network pre-trained on ImageNet~\cite{imagenet}.


\section{Discussion}

\subsection{Differences from Mixup and its derivatives.}

Mixup and its derivatives mix image contents within a random image patch, which does not consider both the spatial as well as semantic information of the image patch carefully. From Table~\ref{tab:ablation_study_for_da}, we find that these random patch can degrade the performance on FID.

In comparison with the traditional Mixup pipelines, OccluMix performs cut-and-paste between fake images and other reference images under semantic guidance, which breaks the defect of random patch mix.


\begin{figure*}[t]
	\begin{center}
		\includegraphics[width=0.98\linewidth]{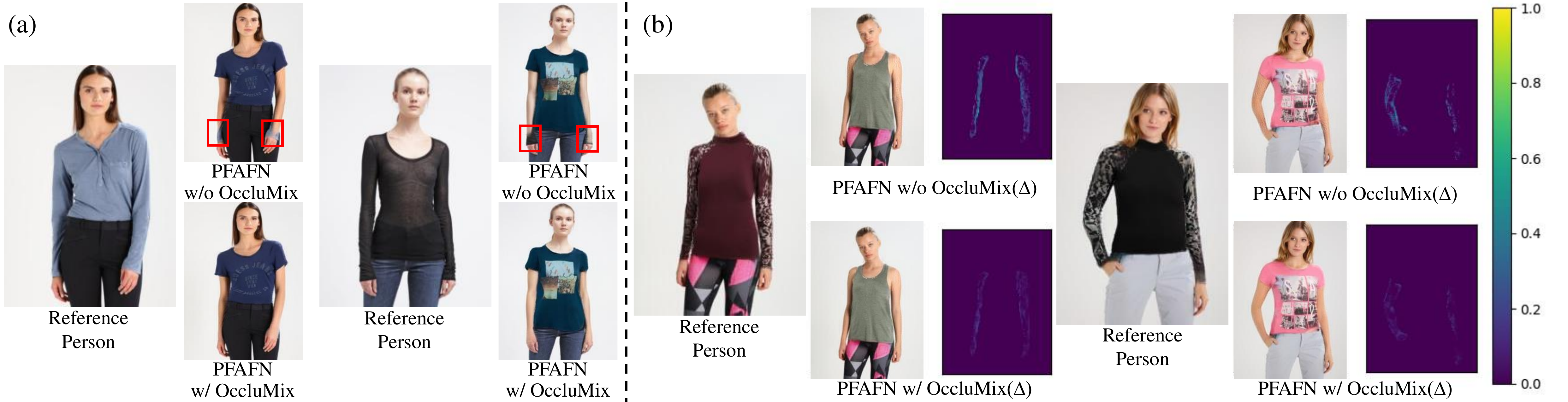}
	\end{center}

	\caption{Qualitative effects of OccluMix during virtual try-on inference. (a) OccluMix successfully generates the clean try-on images while the baseline generates the try-on images with the occlusion from previous clothes. (b) $\triangle$ is the absolute residual intensity map between the try-on image and the ground truth. From the comparison on $\triangle$, it can be demonstrated that the OccluMix resolves the ghost of clothing texture. In both (a) and (b), OccluMix generates the try-on images with better visual coherence. Zooming up for better view. }
	\label{fig:residual_map}

\end{figure*}

\subsection{What does the model learn with Occlusion Mixup?}

\begin{table}[t]
	\centering
	  \caption{FID comparison with mixup and its derivatives. We report the baseline model (PF-AFN~\cite{PFAFN}) performance that is trained on the VITON\cite{Viton} dataset. $\delta$ denotes the performance gap between with and without augmentation.}
	      \vspace{4mm}
        \begin{tabular}{c|cc}
        \hline
        Method                    & FID~$\downarrow$ ($\delta$)  \\ \hline
        
        PF-AFN & 10.09 (+0.00)   \\ \hline
        Cutout~\cite{cutout} & 10.36 (-0.27)                     \\ 
        CutMix~\cite{cutmix}  & 10.31 (-0.22)                   \\ 
        Mixup~\cite{mixup}    & 10.17 (-0.08)                   \\
        CutBlur~\cite{cutblur}  & 9.94 (+0.15)                \\ \hline
        OccluMix  & 9.66 (+0.43)                \\ \hline
        \end{tabular}
\label{tab:ablation_study_for_da}
\end{table}

        
Similar to the other DA methods that prevent the models from making a prediction over-confidently, OccluMix prevents the model from having no distinction between simple images or distorted images and helps it to alleviate the occlusion effect. This can be demonstrated in Fig.~\ref{fig:residual_map}. Note that the superiority of OccluMix is not only reflected in the visual coherence of the try-on images, but also in the decreasing of the residual intensity map. We hypothesize that this enhancement is due to the balanced distribution of fake images. Now the model has learned to distinguish between simple and distorted data, and this leads the model to learn "where" and "how" it should address the occlusion effects.

\section{Experiments}
In this section, we first compare the DOC-VTON with state-of-the-art methods. Then, a detailed ablation study is made to analyze each component. Finally, we provide massive occlusion samples to verify Inherent- and Acquired-Occlusion and further demonstrate the negative effects of occlusion on the human visual system. Besides, to further validate the performance of the proposed DOC-VTON, we perform the visual comparison of the occlusion images (Generated by PF-AFN) and clean images (Generated by our
DOC-VTON).

\subsection{Dataset}
Experiments are conducted on the VITON dataset~\cite{Viton} that used in CP-VTON~\cite{cp-Vton}, CP-VTON+~\cite{CP_VTON_plus}, ACGPN~\cite{ACGPN} and PF-AFN~\cite{PFAFN}. VITON contains a training set of 14221 image pairs and a testing set of 2032 image pairs, each of which has a target clothing image and woman photo with the resolution of 256 $\times$ 192. VITON-HD~\cite{VITON_HD} consists of 11,647 groups for training set and 2032 groups for testing set, each image with the resolution of 1024 $\times$ 768. All of our evaluations and visualizations are performed on the testing set.
\subsection{Implementation Details}

\subsubsection{Architecture}
DOC-VTON contains CWM, OccluMix and GM. The structure of CWM consists of a dual pyramid feature extraction
network (PFEN)~\cite{PFEN} and a progressive appearance flow estimation
network (AFEN)~\cite{PFAFN}. The generators of Sharpened Parsing Network in OccluMix have the same
structure of U-Net~\cite{Unet}. And the structure of GM is Res-UNet~\cite{clothflow}.
In our experiments, the resolution of images is 256 $\times$ 192 for VITON, and 1024 $\times$ 768 for VITON-HD dataset.

\begin{figure*}[t]
	\begin{center}
		\includegraphics[width=0.98\linewidth]{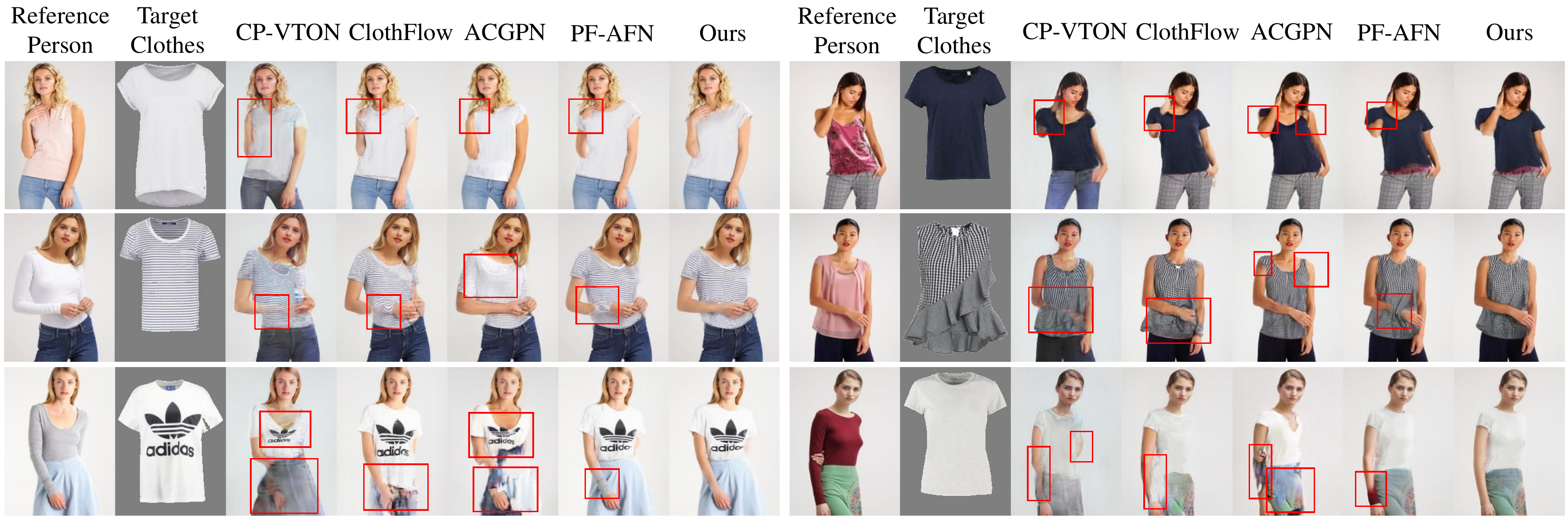}
	\end{center}
	\vspace{-3mm}
	\caption{Visual comparison of VITON dataset. Compared with four state-of-the-art try-on methods~\cite{Viton,ACGPN,PFAFN,cp-Vton}, our model generates more realistic try-on images. \textbf{With the proposed De-occlusion strategy, our approach not only processes the irrational parts of the warping clothes, but also clearly avoids the ghost of former clothes.}}
	\label{fig:Qualitative_Results}
	\vspace{-3mm}
\end{figure*}

\begin{figure}[h]
	\includegraphics[width=0.78\linewidth]{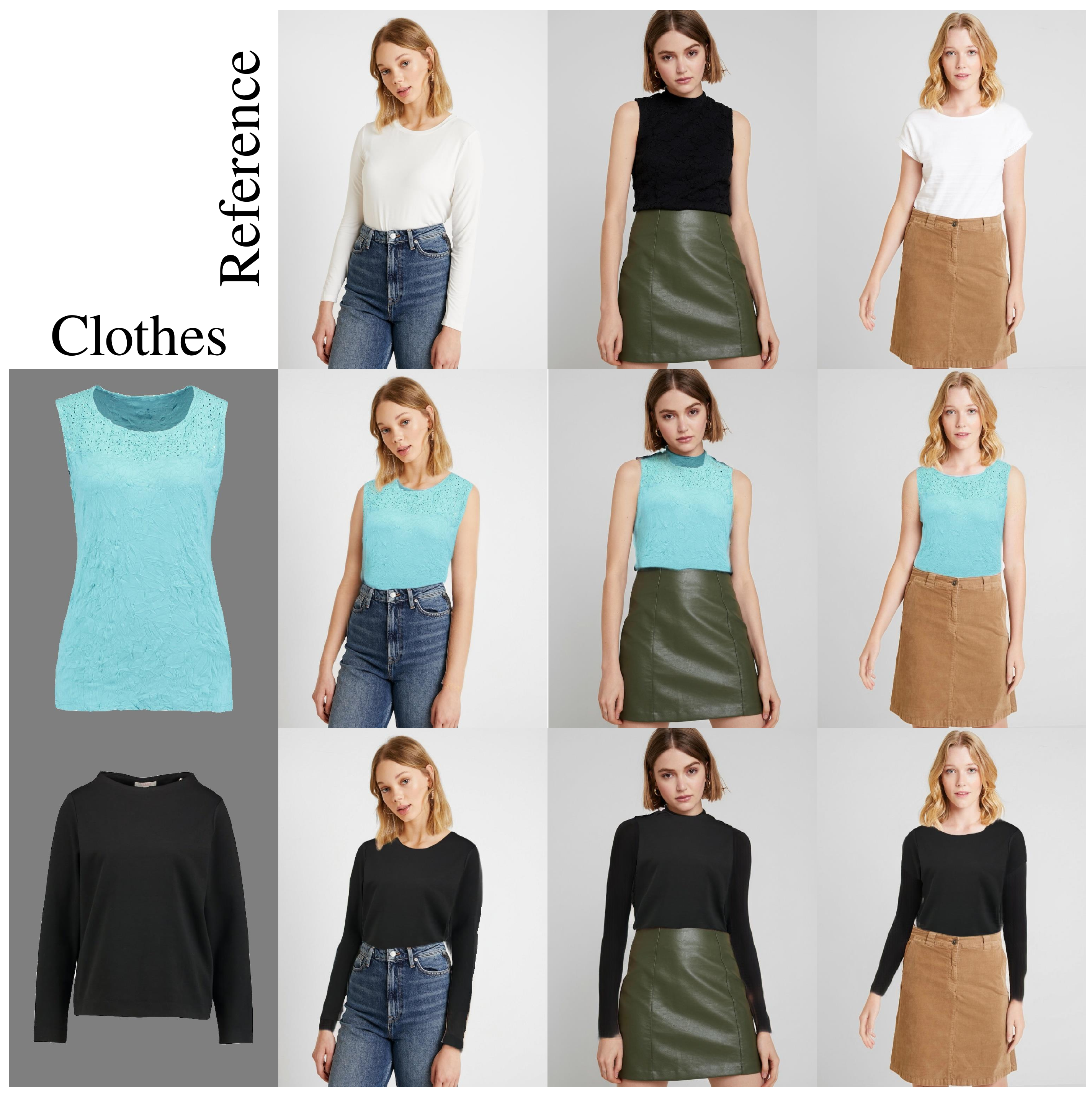}
	\vspace{3mm}
	\caption{Extensive visual comparisons on VITON-HD dataset. From diverse perspectives, our approach generates high-quality images.}
	\label{fig:Viton_hd}
	\vspace{-6mm}
\end{figure}
\subsubsection{Training}We train Sharpened Parsing Generator ($G$) and Restorer ($R$) with 200 epochs. Due to the large spatial transformation between arms and clothes, we use a hard-example mining strategy. In the first 100 training epochs, we selectively train our model with challenging body parts (e.g., arms). In the last 100 epochs, we train our model on the whole body parts (including head, arms and pants) together. Then, we use OccluMix to generate samples for GM. All of them use the initial learning rate as 0.0005 and the network is optimized by Adam optimizer with the hyper-parameter $\beta_1$ = 0.5, and $\beta_2$ = 0.999.

\subsubsection{Testing}
The testing process follows the same procedure
as training. The reference person images, target clothes, human
parsing results, and human pose estimations are given as input
of DOC-VTON to generate the try-on image.

\subsection{Qualitative Results}
To further validate the performance of the proposed DOC-VTON framework, we perform the visual comparison of our proposed method with CP-VTON~\cite{cp-Vton}, ClothFlow~\cite{clothflow}, ACGPN~\cite{ACGPN} and PF-AFN~\cite{PFAFN}. 
\subsubsection{Results on VITON}

As shown in the first and the second rows of Fig.~\ref{fig:Qualitative_Results}, when a person strikes a complex posture, such as standing with one arm raised around the face, the occlusion of target cloth occurs on the arms. In such cases, baseline models all fail to handle the warping process, leading to distorted arm images or broken sleeves. The warping methods of the baseline model fail to restore the severe non-rigid deformation, due to the limited degrees of freedom in TPS~\cite{tps} or misalignment in the Appearance Flow~\cite{clothflow}.

In the last row, images generated by baseline methods contain the obvious artifacts, as CP-VTON~\cite{cp-Vton}, ClothFlow~\cite{clothflow} and ACGPN~\cite{ACGPN} own cluttered texture, boundary-blurring and color mixture problems. To this end, they are vulnerable to segmentation errors as they heavily rely on parsing results to drive image generation. Although PF-AFN~\cite{PFAFN} drives image generation without using parsing results, it has not performed better on generating the fake body parts. Furthermore, when a huge displacement exists between the target clothes and the original clothes (e.g., the person wears long sleeve clothes while the target clothes are short sleeves), PF-AFN~\cite{PFAFN} fails to generate arms at the cuffs of long sleeves since the model does not understand the shape of the clothes thoroughly. To this end, state-of-the-art models are less robust to handle some special cases (e.g., cuffs of long sleeves, complex clothes texture). 

In comparison, the proposed DOC-VTON performs a realistic virtual try-on, which simultaneously handles the warping process to avoid the pernicious occlusion, and preserves the details of both target clothes and human body parts. Benefiting from the OccluMix, our model is robust to generate non-label body parts (e.g., arms, hands, and fingers) from the original cloth. All qualitative results of DOC-VTON clearly verify the superiority against CP-VTON, ClothFlow, ACGPN and PF-AFN.

\subsubsection{Results on VITON-HD} To verify the generalization of our method, we visualize the results on VITON-HD~\cite{VITON_HD} dataset. Since the artifacts in higher resolution are more obvious to be observed, it is more challenging to generate highly-realistic try-on results. As shown in Fig.~\ref{fig:Viton_hd}, the results demonstrate that our DOC-VTON is effective to generate high-quality images on VITON-HD dataset.

\subsection{Quantitative Evaluation}

For virtual try-on, the try-on image is generated by a target cloth and a reference person image.
Since the ground truth of the try-on image is absent, we can not use point-to-point indicators (e.g., SSIM, PSNR and LPIPS) that require computation with labels. In this paper, we adopt the Frechet Inception Distance (FID)~\cite{fid} to measure the diversity of the try-on images. The lower score of FID indicates a higher quality of the results. Besides, the Inception Score (IS)~\cite{is} will not be used, owing to Rosca et.al~\cite{rosca2017variational} have pointed out that applying the IS to the models trained on datasets other than ImageNet will give misleading results.

\begin{table*}[t]
\centering
\caption{The FID score among different methods on the VITON dataset.}
\setlength{\tabcolsep}{5mm}{
\begin{tabular}{c|ccc}
\hline
\multirow{2}{*}{Method} & \multicolumn{1}{c|}{Region of Arm} & \multicolumn{1}{c|}{Warped Clothes} & Try-on Results \\ \cline{2-4} 
                        & \multicolumn{3}{c}{FID~$\downarrow$}                                                                   \\ \hline
CP-VTON                 & \multicolumn{1}{c|}{24.42}         & \multicolumn{1}{c|}{33.17}          & 24.43          \\
CP-VTON+                & \multicolumn{1}{c|}{22.36}         & \multicolumn{1}{c|}{30.21}          & 21.08          \\
ACGPN                   & \multicolumn{1}{c|}{16.62}         & \multicolumn{1}{c|}{24.95}          & 15.67          \\
ClothFlow               & \multicolumn{1}{c|}{18.42}         & \multicolumn{1}{c|}{22.50}          & 14.43          \\
{\color{blue}DCTON}                  & \multicolumn{1}{c|}{{\color{blue}13.27}}         & \multicolumn{1}{c|}{{\color{blue}30.18}}          & {\color{blue}14.82}         \\ 
{\color{blue}RT-VTON}                  & \multicolumn{1}{c|}{{\color{blue}12.70}}         & \multicolumn{1}{c|}{{\color{blue}20.59}}          & {\color{blue}11.66}         \\ 
PF-AFN                  & \multicolumn{1}{c|}{12.86}         & \multicolumn{1}{c|}{18.67}          & 10.09          \\ \hline
Ours                    & \multicolumn{1}{c|}{\color{red}\textbf{11.14}}         & \multicolumn{1}{c|}{\color{red}\textbf{18.18}}         & {\color{red}\textbf{9.54}}           \\ \hline
\end{tabular}}
\label{tab:FID_score}
\end{table*}

Table~\ref{tab:FID_score} lists the FID score of the try-on results for CP-VTON~\cite{cp-Vton}, CP-VTON+~\cite{CP_VTON_plus}, ClothFlow~\cite{clothflow}, ACGPN~\cite{ACGPN}, DCTON~\cite{DCTON}, RT-VTON~\cite{RT_VTON}, PF-AFN~\cite{PFAFN} and our DOC-VTON on the VITON dataset. Our proposed DOC-VTON outperforms other methods, which indicates that DOC-VTON can improve the perceptual quality of try-on images, handle large misalignment between clothes and person, and synthesize realistic try-on results. 


\begin{figure}[t]
\centering
\includegraphics[width=0.48\linewidth]{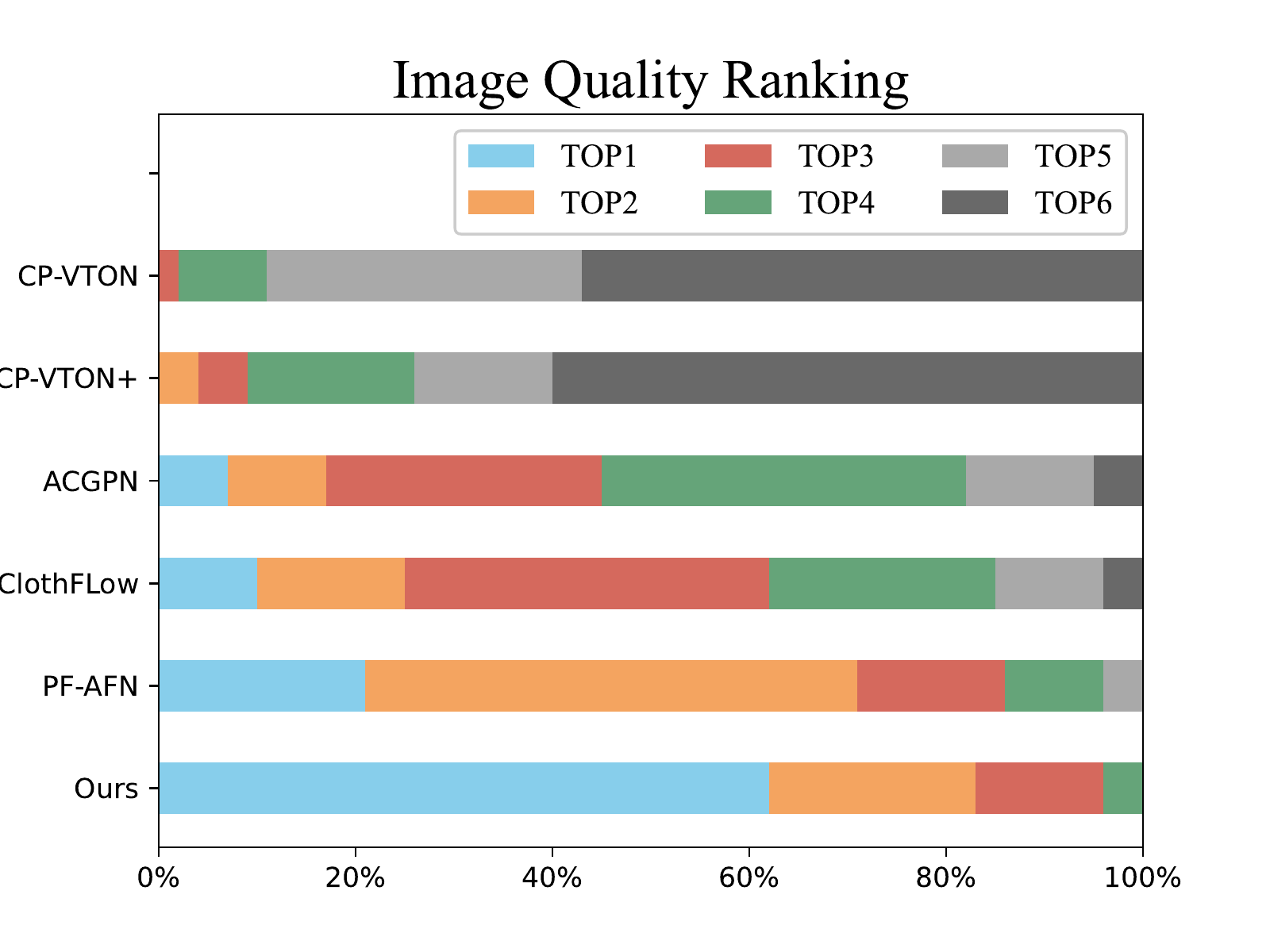}
\includegraphics[width=0.48\linewidth]{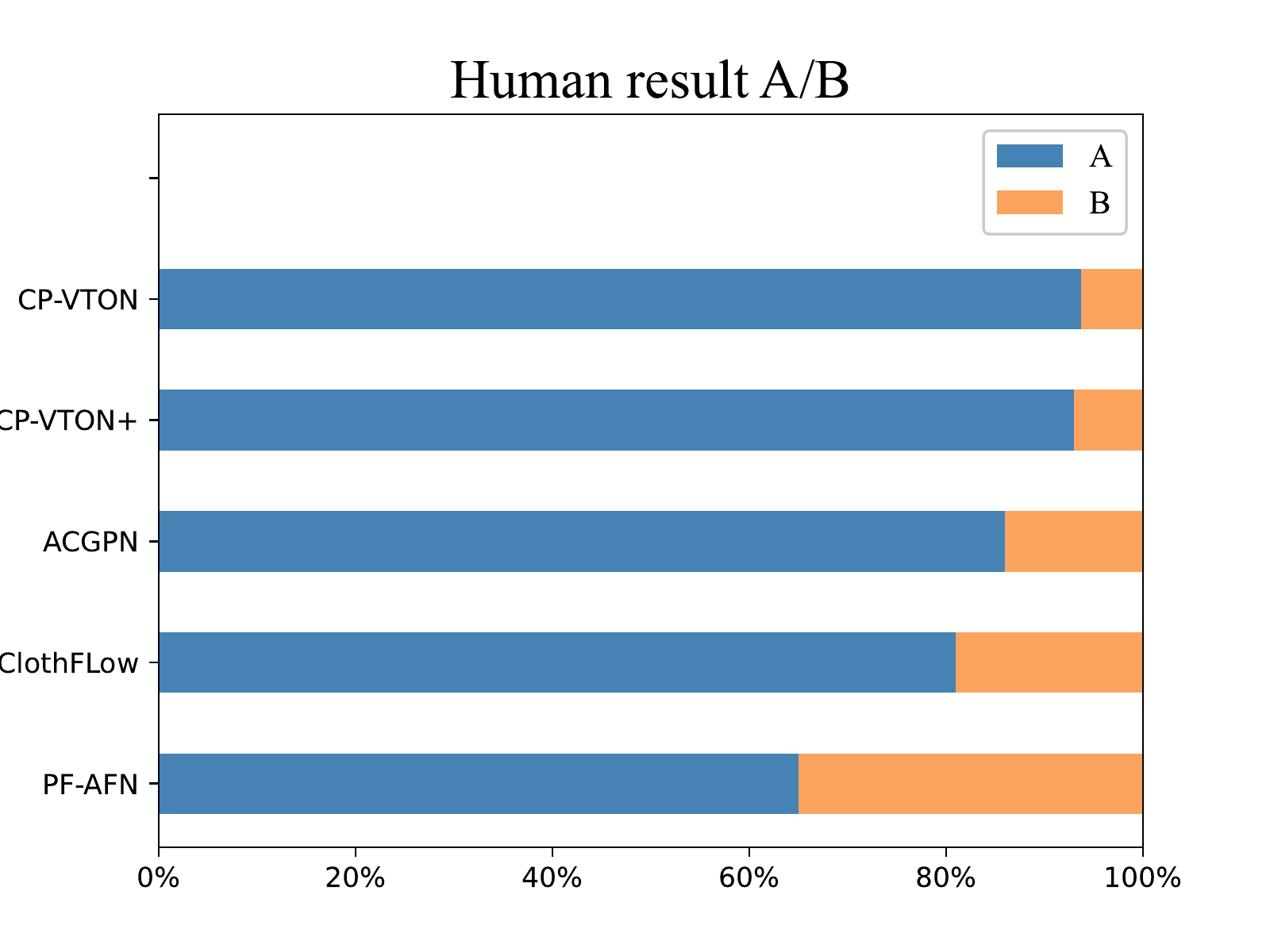}
\vspace{4mm}
\caption{We make an in-depth quality assessment on user study by recruiting 25 volunteers to complete two tasks. The left image is the quality ranking of each method. The right image is the comparison between DOC-VTON and other methods by a A/B test. In the right image, the label A denotes the percentage where our DOC-VTON is considered better over the compared method, and the label b denotes the percentage where the compared method is considered better over our DOC-VTON.}
\label{fig:user_study}
\vspace{-6mm}
\end{figure}

\subsection{Human Perception Study}

The subtle changes (e.g. removing partial occlusion or ghost) on try-on images play an important role in human visual coherence. Because FID is insensitive to subtle changes on synthetic images, which can not demonstrate the effectiveness of our method. We further conduct two user study by recruiting 25 volunteers. For the first task, our goal is to verify the superiority of the individual methods. Specifically, CP-VTON~\cite{cp-Vton}, ClothFlow~\cite{clothflow}, ACGPN~\cite{ACGPN}, PF-AFN~\cite{PFAFN}, and DOC-VTON generate the try-on images from 300 specified reference images respectively. Each volunteer is asked to rank the methods in each group images. The left image of Fig.~\ref{fig:user_study} demonstrate a clear advantage. Our method ranks the first 61.83 $\%$. The second task is to compare our method with other methods one by one. We divide the previous synthetic images into five a / b (\textit{i.e.,} a is our methods and b is other methods) groups. Each volunteer is asked to choose the one with better visual quality. As shown in the right image of Fig.~\ref{fig:user_study}, our DOC-VTON is always rated better than the other methods.

As shown in Fig.~\ref{fig:user_study}, our DOC-VTON achieves the highest voter turnout in both tasks. It verifies the great superiority of DOC-VTON over the other methods. The human perception study demonstrates the effectiveness of the proposed method in removing pernicious occlusions and enjoying the human visual system in the try-on task.

\subsection{Ablations}
In this section, we analyze each component in OccluMix which brings to the robustness of the model. 

\begin{figure}[t]
\centering
\includegraphics[width=0.48\linewidth]{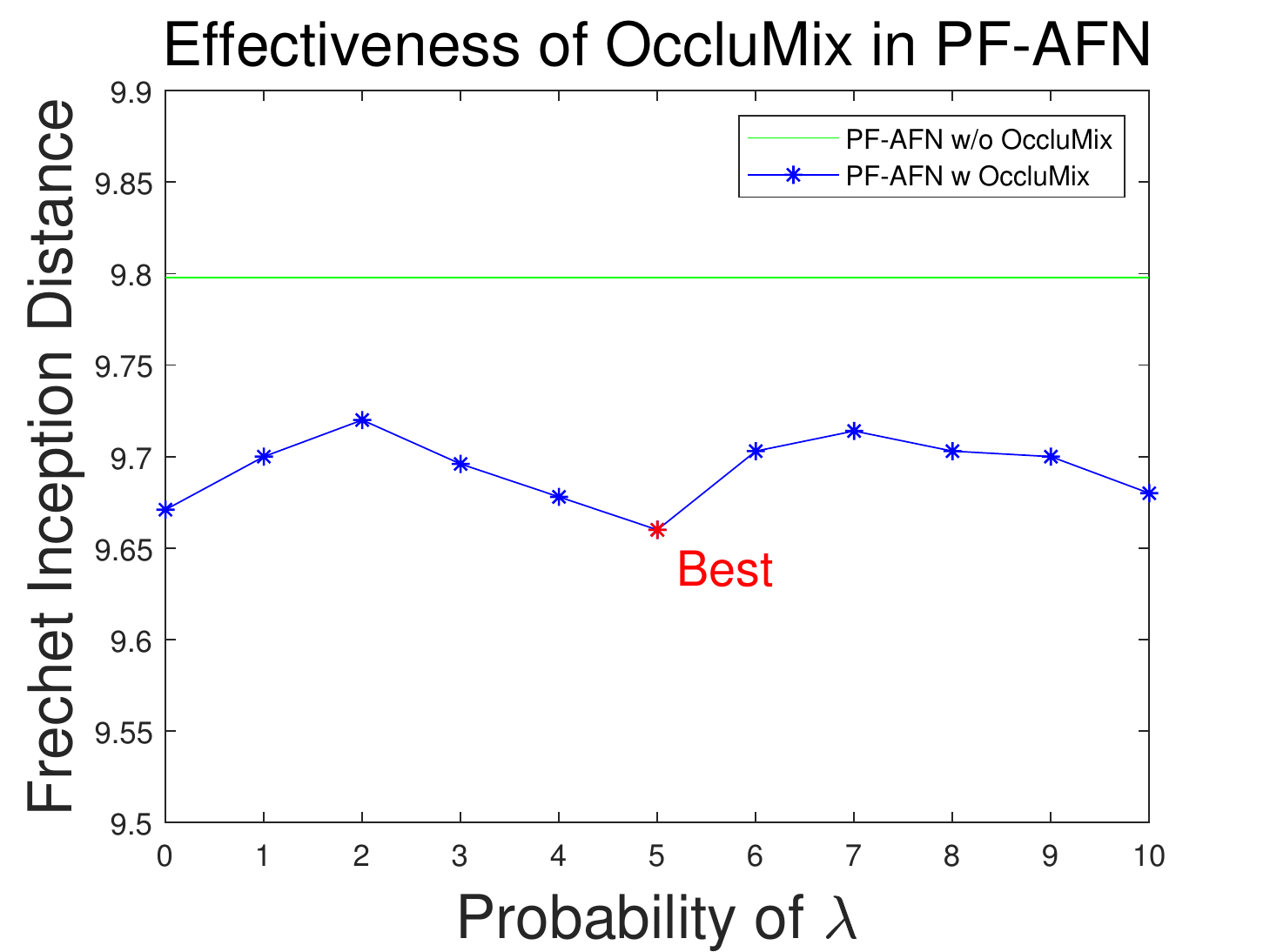}
\includegraphics[width=0.48\linewidth]{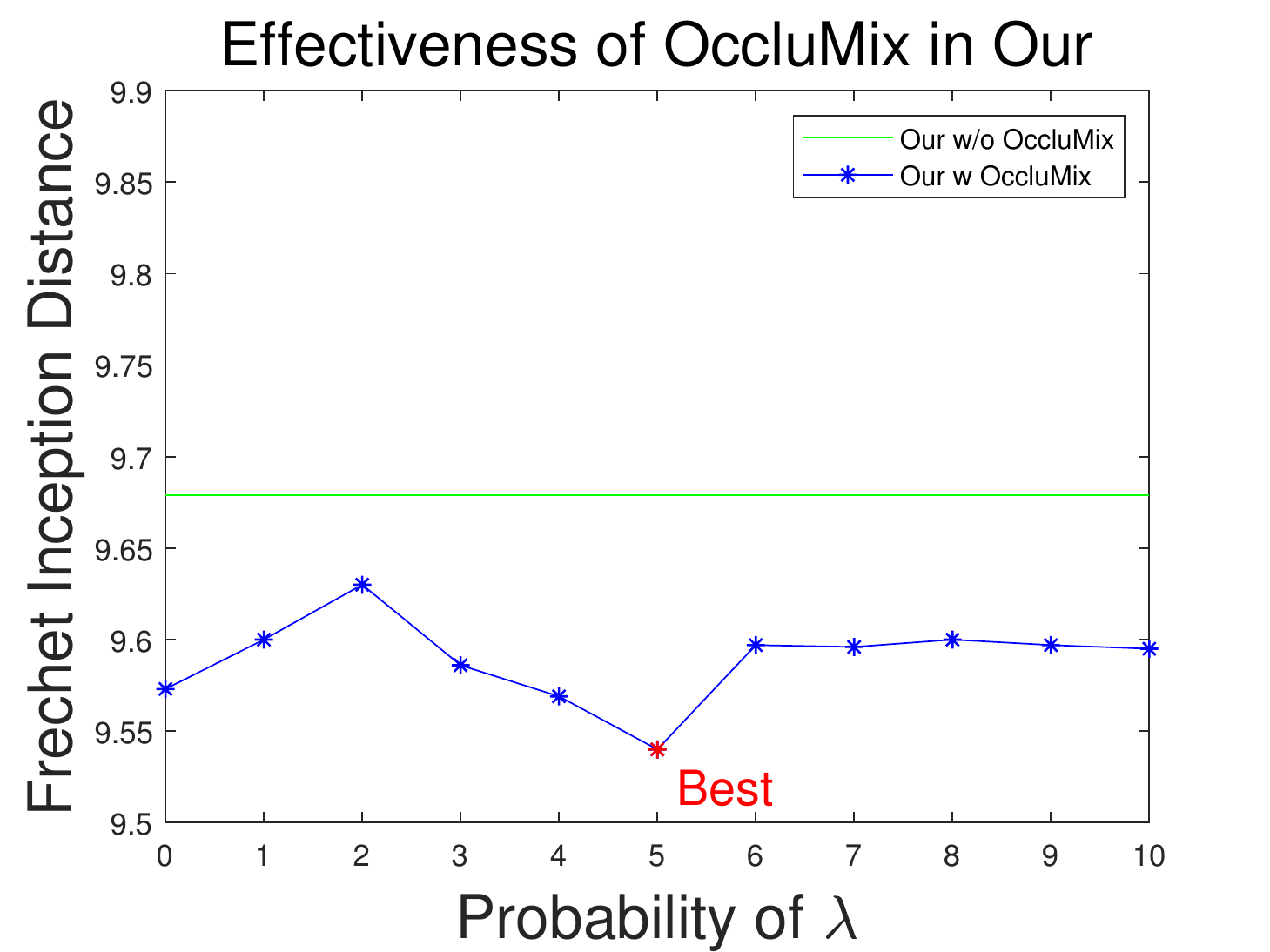}
\caption{Ablation study of complex coefficient \texorpdfstring{$\lambda$}{Lg}.}
\label{fig:ablation_study_3}
\vspace{-3mm}
\end{figure}

\textbf{\textit{Analysis of Complex Coefficient \texorpdfstring{$\lambda$}}}.

As discussed above, we introduce the OccluMix scheme in the training stage. To validate the effectiveness of OccluMix, we design a baseline method (\textit{i.e.,} `w/o OccluMix') that is finetuned on the original model. As shown in Fig.~\ref{fig:ablation_study_3}, `w/o OccluMix' meets a 0.08 FID drop when removing OccluMix scheme. To analyze the OccluMix scheme continuously, we conduct experiments by setting $\lambda$ from 0 to 1, with intervals of 0.1. The continuous sampling of {$\lambda$} indicates the mixup paradigm from simple to complex. As shown in Fig.~\ref{fig:ablation_study_3}, we find that the model shows the trivial result when the simplest mixup pattern is used ($\lambda = 0.0$ indicates the DOC-VTON only use simple textures for OccluMix). We also conduct a balance distribution with $\lambda = 0.5$, which indicates two well-proportioned texture complexity are used for OccluMix. As shown in Fig.~\ref{fig:ablation_study_3}, the balance distribution shows the best performance than others. Since the unbalanced mixup distribution lead to the performance suffering from an obvious drop, we adopt a balanced complex coefficient in OccluMix.

\begin{table}[t]
	\centering
	\vspace{2mm}
	  \caption{Ablation studies of the DOC-VTON. Lower FID indicates better results.}
	      \vspace{0mm}
        \begin{tabular}{c|ll}
        \hline
        Method                    & FID~$\downarrow$  \\ \hline
        
        w/o SPN & 9.80   \\
        w/ SPN & 9.68                     \\ 
        w/o OccluMix    & 9.80                   \\ 
        w/ OccluMix    & 9.66                   \\ \hline
        Ours           & 9.54    \\ \hline
\end{tabular}
\label{ablation_study_detail}
\end{table}


        



\textbf{\textit{Ablations of OccluMix.}}

We show the ablation studies on the effects of using the OccluMix scheme in the training stage. Meanwhile, the `w/o OccluMix' owns the same network architecture as `w/ OccluMix', other than the training process abandons the OccluMix strategy and SPN. As shown in Fig.~\ref{fig:residual_map}, we notice that the try-on image generated by the model trained with the OccluMix scheme outperforms the plain model. With the OccluMix, the model enables to learn where to generate realistic body parts. As depicted in Table~\ref{ablation_study_detail}, `w/ OccluMix' achieves a 0.14 FID gains compared with `w/o OccluMix'.

\textbf{\textit{Sharpened Parsing Network.}}

\begin{figure}[ht]
	\begin{center}
		\includegraphics[width=0.98\textwidth]{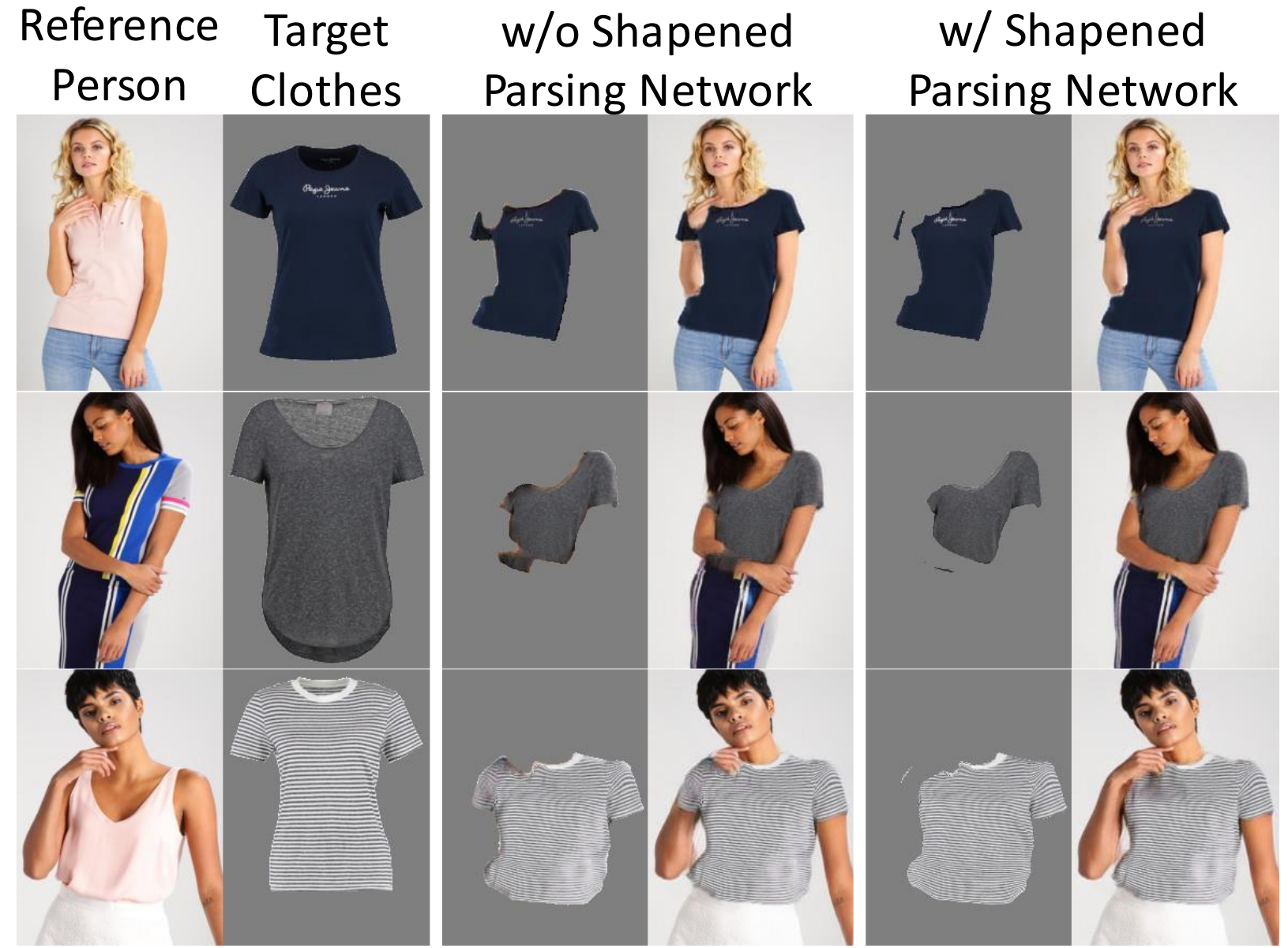}
	\end{center}
	\vspace{0mm}
	\caption{Ablation studies on the effects of the Sharpened Parsing Network.}
	\label{ablation_SPG}
	\vspace{-3mm}
\end{figure}

We show the ablation studies on the effects of the `Sharpened Parsing Network (SPN)'. Since the result of SPN includes semantic information about the body parts of a try-on image, we can apply them to guide the warping process when the target clothes produce unreasonable distortion on the human body (e.g., arms, hands). As shown in Fig.~\ref{ablation_SPG}, with the guidance of the SPN, the warping clothes overcome the complex spatial interactions, and the distortion between the rough warping clothes and the body part. Otherwise, the rough shape of warping clothes will cause body distortion. As depicted in Table~\ref{ablation_study_detail}, `w/ SPN' achieves a 0.12 FID gains compared with `w/o SPN'.

\section{Conclusion and limitation}
\label{section: Conclusion}
In this paper, we have introduced Occlusion Mixup (OccluMix), a new DA method and strategy for training a stronger try-on model. We proposed a novel De-Occlusion approach by a data augmentation manner, which enables our model to remove the partial occlusion to produce realistic try-on images. Extensive evaluations clearly verify the obvious superiority of DOC-VTON over the state-of-the-art methods with less occlusion effect.

Though DOC-VTON addresses the occlusion problem in specific try-on datasets, it still shows limited performance on out-of-distribution (OOD) images. Thus, DOC-VTON may suffer restrictions of light, posture, and background conditions. In the following work, we will continue to apply unsupervised 2D to 3D transformation into DOC-VTON to develop an ODD virtual try-on framework. In addition, since our method can be used not only to remove occlusions, but also to expose body parts, we have declared BSD license in the open source code to avoid this potential social implications.

\bibliographystyle{IEEEtran}
\bibliography{egbib}
\end{document}